\documentclass[10pt,twocolumn,letterpaper]{article}
\usepackage{authblk}

\usepackage[pagenumbers]{cvpr}              %

\renewcommand{\paragraph}[1]{\smallskip\noindent\textbf{#1}}

\usepackage{xcolor}

\usepackage{bm}
\usepackage{algorithm}
\usepackage[noend]{algpseudocode}

\usepackage{amsmath}

\definecolor{cvprblue}{rgb}{0.21,0.49,0.74}
\usepackage[pagebackref,breaklinks,colorlinks,allcolors=cvprblue]{hyperref}

\def\paperID{60} %
\def\confName{EarthVision}
\def\confYear{2025}

\newcommand{\dataset}[0]{S-EO}

\title{\dataset{}: A Large-Scale Dataset for Geometry-Aware Shadow Detection \\ in Remote Sensing Applications\vspace{-0.3em}}

\author[1,2]{\textit{El\'ias Masquil}}
\author[3]{ \hspace{1em}\textit{Roger Mar\'i}}
\author[4]{\hspace{1em}\textit{Thibaud Ehret}}
\author[5]{\hspace{1em}\textit{Enric Meinhardt-Llopis}}
\author[1,5]{\\\textit{Pablo Musé}}
\author[5]{\hspace{1em}\textit{Gabriele Facciolo}}

\affil[1]{IIE, Facultad de Ingeniería, Universidad de la República, Uruguay}
\affil[2]{Digital Sense, Uruguay}
\affil[3]{Eurecat, Centre Tecnològic de Catalunya, Multimedia Technologies, Barcelona, Spain}
\affil[4]{AMIAD, Pôle Recherche, France}
\affil[5]{Université Paris-Saclay, ENS Paris-Saclay, CNRS, Centre Borelli, 91190, Gif-sur-Yvette, France}

\begin{document}
\maketitle
\captionsetup{skip=6pt}

\begin{abstract}
 We introduce the \dataset{} dataset: a large-scale, high-resolution dataset, designed to advance geometry-aware shadow detection. Collected from diverse public-domain sources, including challenge datasets and government providers such as USGS, our dataset comprises $702$ georeferenced tiles across the USA, each covering $500 \times 500$~m. Each tile includes multi-date, multi-angle WorldView-3 pansharpened RGB images, panchromatic images, and a ground-truth DSM of the area obtained from LiDAR scans. For each image, we provide a shadow mask derived from geometry and sun position, a vegetation mask based on the NDVI index, and a bundle-adjusted RPC model. With approximately 20,000 images, the \dataset{} dataset establishes a new public resource for shadow detection in remote sensing imagery and its applications to 3D reconstruction.
 To demonstrate the dataset's impact, we train and evaluate a shadow detector, showcasing its ability to generalize, even to aerial images. Finally, we extend EO-NeRF~—~a state-of-the-art NeRF approach for satellite imagery~—~to leverage our shadow predictions for improved 3D reconstructions.

\end{abstract}
    
\vspace{-2em}

\vspace{1em}
\section{Introduction}
\label{sec:intro}

Shadows play a crucial role in visual perception, providing key insights into depth, contours, textures, and lighting within a 3D scene. Although they can hide scene details by darkening regions where light is obstructed, they also encode valuable information about object shapes~\cite{irvin1989methods,liasis2016satellite} and spatial relationships, enabling scene interpretation without requiring additional data sources. Whether the goal is to remove shadows to recover hidden details or to leverage them as geometric cues, accurate shadow detection is an essential task for image-based scene understanding.

\begin{figure}
    \centering
    \begin{tabular}{@{\hskip 0cm}c@{\hskip 0.1cm}c@{\hskip 0.1cm}c@{\hskip 0.1cm}c}
    \includegraphics[width=0.24\linewidth]{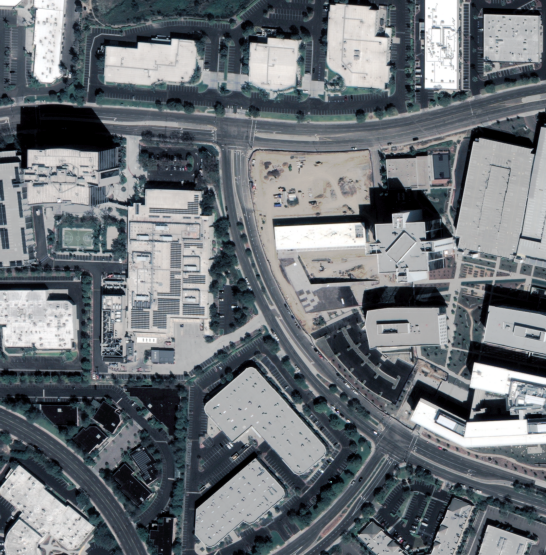} &
    \includegraphics[width=0.24\linewidth]{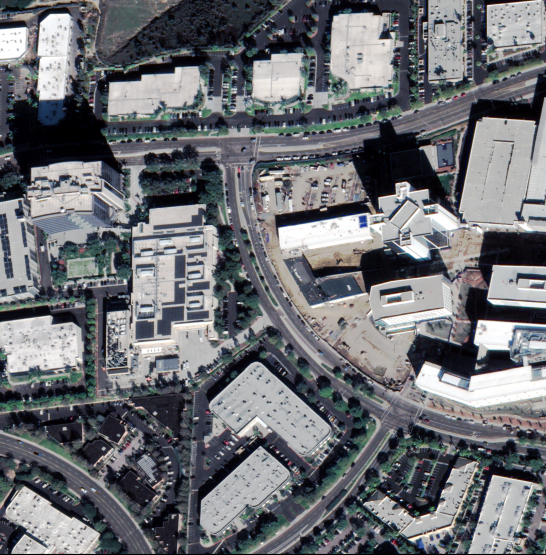} &
    \includegraphics[width=0.24\linewidth]{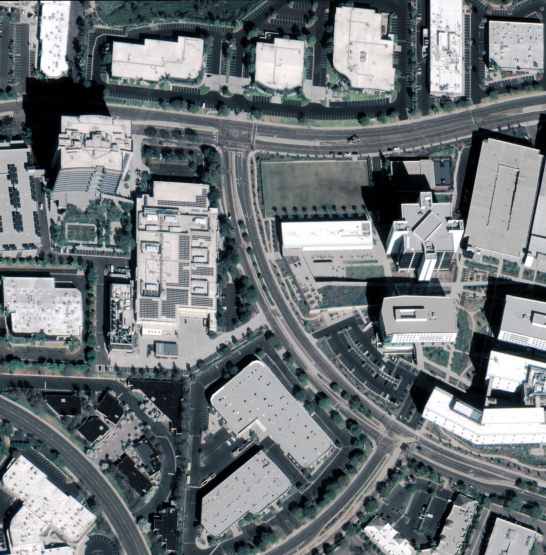} &
    \includegraphics[width=0.24\linewidth]{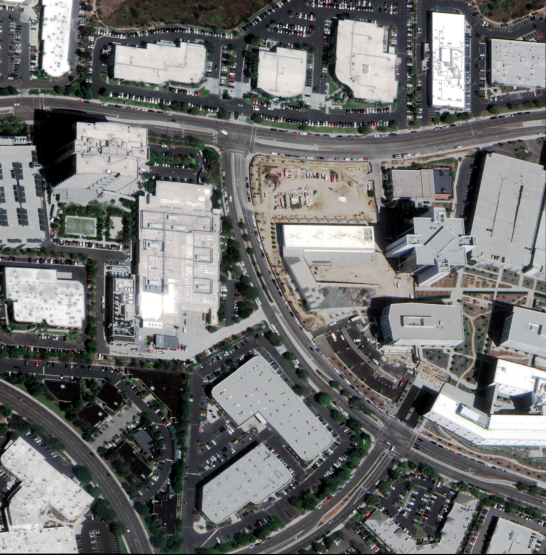} \\
    \includegraphics[width=0.24\linewidth]{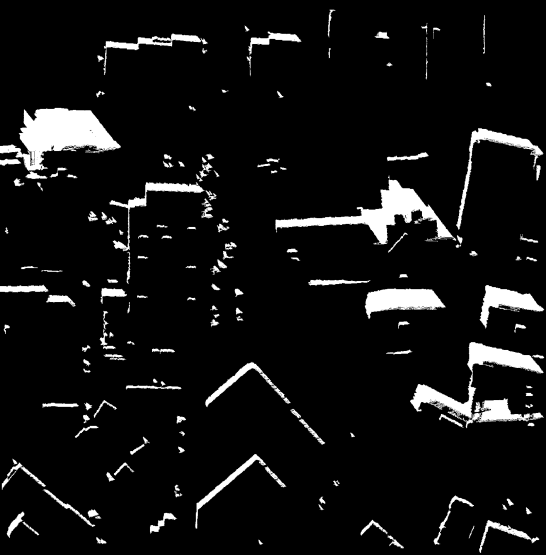} &
    \includegraphics[width=0.24\linewidth]{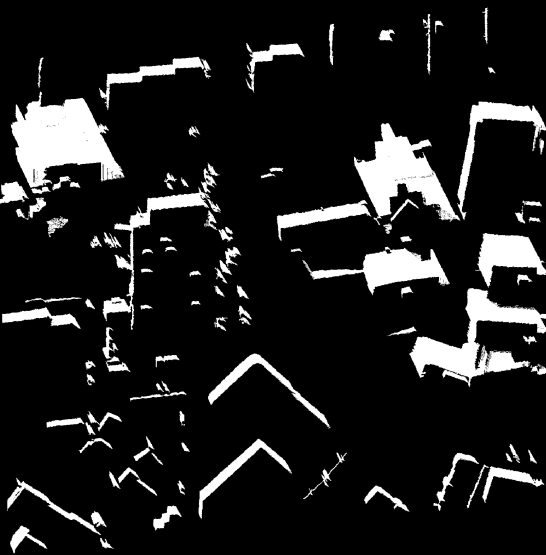} &
    \includegraphics[width=0.24\linewidth]{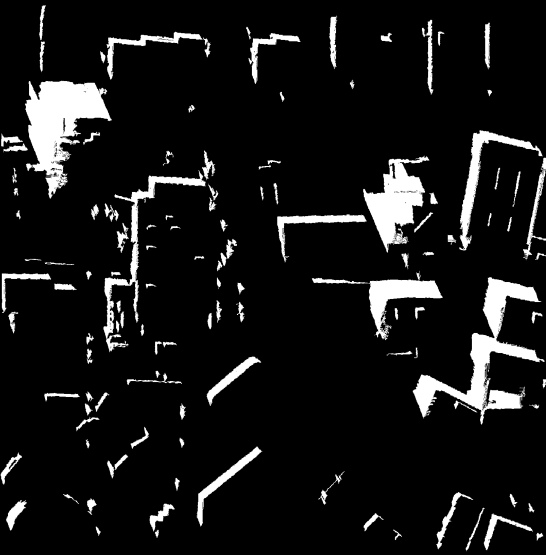} &
    \includegraphics[width=0.24\linewidth]{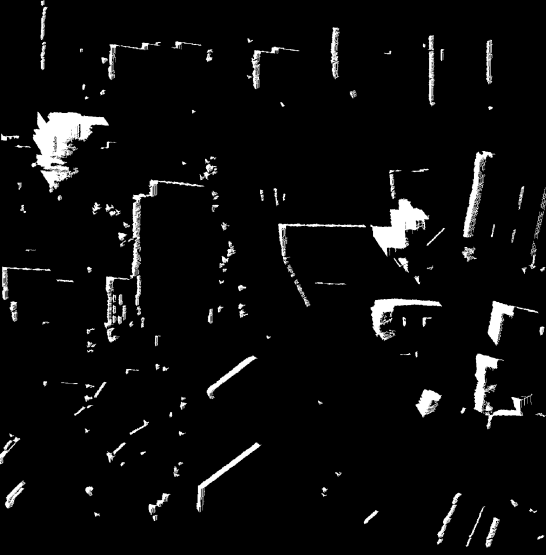}
    \end{tabular}
    
    \caption{Example multi-date satellite images and shadow masks of an area from the \dataset{} dataset. Shadows and their variations are critical cues for Earth Observation algorithms focused on geometry estimation and scene understanding.}
    \label{fig:teaser}
\end{figure}

As shadows are present in nearly all satellite images~\cite{AISD}, the development of accurate shadow detection methods is particularly important to create efficient systems dedicated to Earth Observation (EO). As in many other domains, state-of-the-art approaches primarily rely on deep learning techniques~\cite{shadowsReview}, which require large, high-quality annotated datasets for effective training. While remote sensing data availability continues to grow, dedicated shadow datasets remain scarce. To our knowledge, only one prior work has publicly released a small dataset \cite{AISD}, with approximately 500 manually annotated images.

To address this limitation, we introduce a large-scale, high-resolution dataset designed to advance research in shadow detection and 3D reconstruction: the \dataset{} dataset (named after Shadow-aware Earth Observation). Our dataset, derived from the IARPA CORE3D program data~\cite{CORE3D,dfc1,dfc2}, consists of extensive multi-date, multi-angle WorldView-3 imagery at 30 cm resolution, along with georeferenced shadow masks and aligned Digital Surface Models (DSMs) derived from USGS LiDAR scans (Map services and data available from U.S. Geological Survey, National Geospatial Program). Shadow masks, depicted in Figure~\ref{fig:teaser}, were automatically generated from the surface models using a shadow-casting algorithm and the sun position. All images and data are delivered in a ready-to-use format: {\small\url{https://centreborelli.github.io/shadow-eo}}.

With more than 700 georeferenced and annotated tiles of $500 \times 500$~m, totaling approximately 20,000 images, and the associated DSMs, the \dataset{} dataset provides a new public resource for both shadow detection and shadow-aware 3D modeling from remote sensing data.
The \dataset{} dataset includes a wide variety of samples, covering diverse cities with different urban layouts, vegetation types and climate, observed at different seasons and times of the day. For instance, it captures snow-covered landscapes, which introduce additional challenges for shadow detection models.

To demonstrate the relevance of this novel dataset, we train a shadow detection model and show that it generalizes effectively, even to aerial datasets. Our experiments find that the U-Net architecture \cite{unet} remains a strong baseline, performing on par with specialized shadow detection models. Furthermore, we extend the state-of-the-art multi-view 3D modeling framework EO-NeRF~\cite{eonerf} to incorporate shadow mask supervision using the detector predictions. Our results show that shadow supervision consistently improves 3D reconstruction performance, boosting both the accuracy of altitude values and the quality of output shapes.

\begin{figure*}[t]
    \centering
    \includegraphics[width=0.95\textwidth]{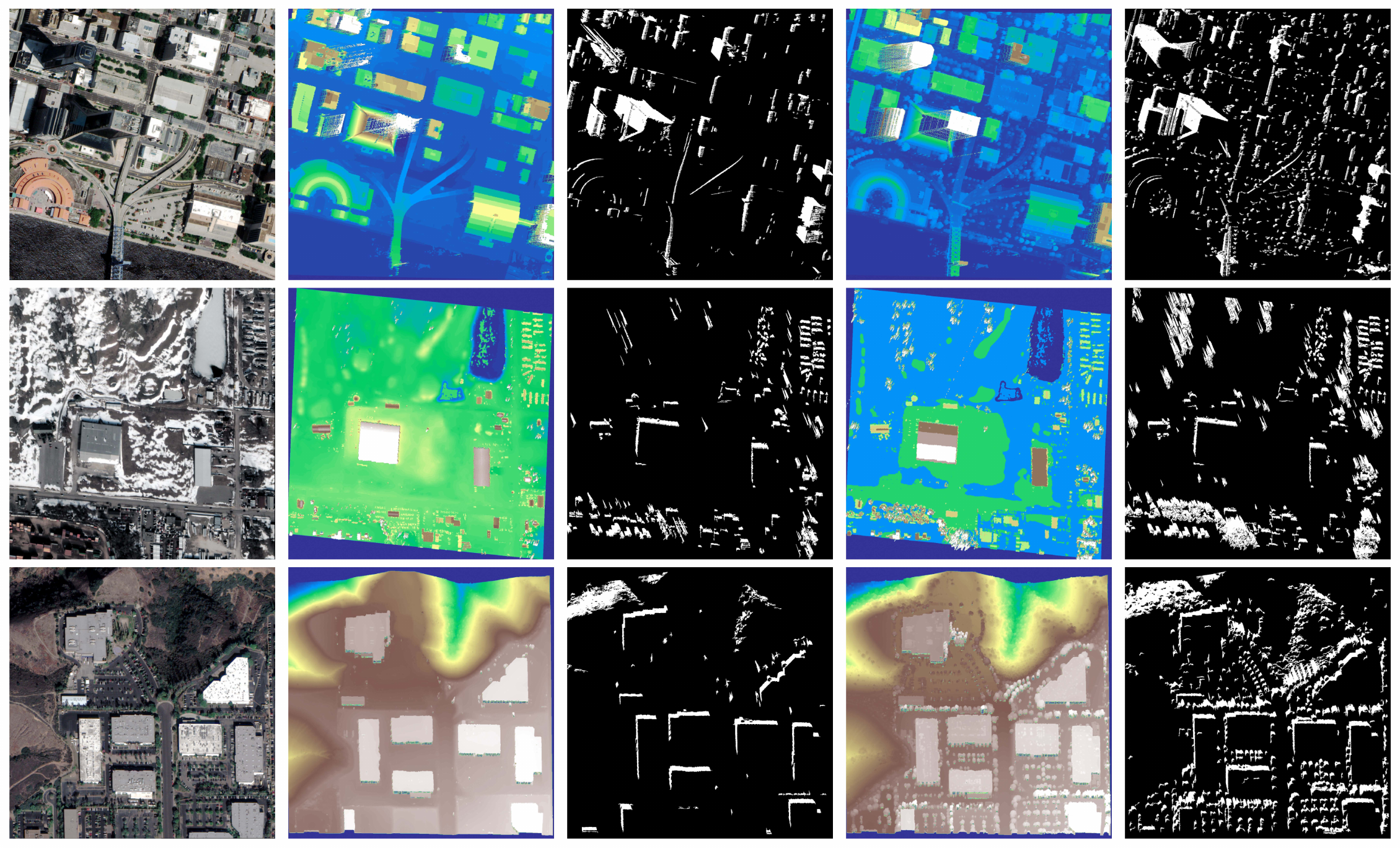}
\caption{Top to bottom: \dataset{} dataset images from three sites— in Jacksonville, Omaha, and San Diego. For each site, left to right: the pansharpened RGB image; the DSM computed from the minimum elevation per grid cell (DSM Min) and its derived shadow mask; and the DSM computed from the maximum elevation per grid cell (DSM Max) with its respective shadow mask. DSM Min yields a cleaner shadow mask by filtering out transient elements (e.g., trees), although it may erode building edges and produce smaller shadows. In contrast, DSM Max preserves more building details and generates larger shadows but with increased noise in the shadow mask.}
    \label{fig:dataset_overview}
\end{figure*}

In summary our contributions consist of:  
\begin{itemize}
    \item[--] \textbf{A large-scale, high-resolution shadow dataset (\dataset{})}, built from public-domain WorldView-3 imagery and USGS LiDAR-derived DSMs. 
    \item[--] \textbf{A shadow detector for remote sensing images}, trained on the \dataset{} data. We evaluate its performance and demonstrate its generalization ability. With minimal fine-tuning, our model surpasses the state-of-the-art in shadow segmentation on previously unseen aerial imagery.  
    \item[--] \textbf{A demonstration of shadow-supervised 3D reconstruction from satellite imagery}. We incorporate shadow-based supervision into EO-NeRF and show that it improves 3D reconstruction quality.
\end{itemize}

Along with the dataset, we release the complete data processing pipeline, our shadow detection model, and our shadow-supervised implementation of EO-NeRF.

\section{Related work}
\label{sec:formatting}

\subsection{Deep supervised shadow detection}
A comprehensive overview of the state-of-the-art in deep shadow detection is provided by \cite{shadowsReview}. Here, we highlight key advances in the field, focusing on both model architectures and datasets.

\paragraph{Models.}
Supervised shadow detection has evolved alongside broader trends in computer vision. Early deep learning approaches~\cite{khan2014automatic, khan2015automatic} replaced hand-crafted features with neural networks, which were later integrated into optimization models such as Conditional Random Fields (CRFs)~\cite{zhu2010learning} for refinement. Over time, the focus shifted to end-to-end models, such as GAN-based approaches~\cite{nguyen2017shadow}, where the shadow detector operates as the generator of a conditional GAN.

Subsequent advancements introduced multi-scale architectures~\cite{hu2018direction} to better capture fine details, while recent state-of-the-art models leverage popular backbones such as ResNeXt~\cite{resnext}, U-Net variants~\cite{unet}, or EfficientNet~\cite{tan2019efficientnet}. Transformers are also being used for shadow detection~\cite{jie2022fast}, to efficiently capture contextual relationships with smaller models and achieve faster inference times.

However, as noted in~\cite{shadowsReview}, comparisons between existing methods often suffer from inconsistencies in input sizes, evaluation metrics, datasets, and implementation details, making it difficult to determine whether newer architectures truly outperform earlier ones.
Since not all models in the literature are publicly available, this work relies on the implementations listed in~\cite{shadowsReview}. The FSDNet~\cite{fsdnet} is selected
as a baseline due to its balance between performance and efficiency. The FSDNet architecture is characterized by Direction-aware Spatial Context (DSC) modules \cite{hu2018direction} and a MobileNet V2 backbone~\cite{sandler2018mobilenetv2}.

\paragraph{Datasets.}
Popular benchmarks for shadow detection typically contain thousands of real-world images from diverse domains. SBU~\cite{sbu}, one of the most widely used datasets, provides 4,087 training images and 638 testing images. However, as noted by~\cite{yang2023silt}, shadow annotations are often noisy and inconsistent. To address this, a method for self-supervised label refinement is introduced for training alongside expert-verified annotations for the test partition.

ISTD~\cite{istd} introduces a dataset designed for both shadow detection and removal, providing triplets of shadow images, shadow masks, and shadow-free images. It contains 1,330 training images, 540 testing images, and 135 distinct backgrounds. CUHK-Shadow~\cite{fsdnet} presents a large-scale dataset of 10,500 shadow images, categorized into different shadow types, including building shadows, Google Maps images, and shadows cast by people and objects.

\subsection{Shadow detection in remote sensing}

AISD~\cite{AISD} was the first publicly available remote sensing dataset for shadow detection. It is derived from the Inria Aerial Image Labeling Dataset~\cite{maggiori2017can} and consists of 514 images with manually annotated shadow masks. The image sizes vary between 256$\times$256 and 1688$\times$1688 pixels. The dataset covers regions in both the United States (Austin, Chicago) and Austria (Tyrol, Vienna, Innsbruck). It is split into 412 training images, while the validation and test sets each contain 51 images with their corresponding masks.

In the same work, the authors introduce DSSDNet, a deep learning architecture for remote sensing shadow detection. The model comprises two key components: (1) an encoder-decoder residual (EDR) structure, which extracts multi-level and discriminative shadow features, and (2) a deeply supervised progressive fusion (DSPF) module, which enhances detection by progressively refining feature maps during training. Unfortunately, the model weights are not publicly available, and the codebase is implemented in MATLAB, making it difficult to use in modern deep learning frameworks.

In \cite{automaticShadows}, the authors address the domain gap in shadow detection models between ground-level images and aerial imagery, as well as the scarcity of annotated satellite image datasets. They propose a pipeline for automatically generating shadow masks using dense 3D point clouds reconstructed from city-scale Wide Area Motion Imagery (WAMI) sequences and the sun position. While their approach shares similarities with ours, there are some key differences. Unlike our method, which leverages large-scale ground-truth 3D models and satellite imagery, their approach relies on 3D point clouds reconstructed from aerial imagery. Their contribution is limited to preliminary experiments and does not include a publicly released dataset.

\subsection{Shadows in advanced 3D modeling}

In recent years, multi-view 3D scene modeling has reached unprecedented levels of photorealism, driven by advanced volumetric representations such as NeRF~\cite{mildenhall2021nerf} and Gaussian Splatting~\cite{kerbl20233d}. The variants of these methods adapted to satellite imagery have emphasized the importance of shadows in accurately interpreting both scene appearance and geometry. Notable examples include Shadow-NeRF~\cite{derksen2021shadow}, Sat-NeRF~\cite{mari2022satnerf}, EO-NeRF~\cite{eonerf}, SUNDIAL~\cite{behari2024sundial}, Sat-NGP~\cite{billouard2024sat}, and EOGS~\cite{aira2024eogs}, among others. These works highlight the importance of incorporating shadow estimation into the optimization process to produce more accurate geometric reconstructions, as shadows and scene geometry are intrinsically linked. Early models predicted shadows directly using embedding vectors based on the sun position, whereas more recent approaches use additional constraints, such as mathematical modeling of light transmittance for analytical shadow computation~\cite{eonerf} and irradiance models of higher complexity~\cite{behari2024sundial}.

Shadow masks have been successfully used for geometry recovery in NeRF-based methods for conventional imagery~\cite{tiwary2022towards, ling2023shadowneus}. With the emergence of new datasets and improved shadow detection techniques for remote sensing images, shadow masks could also be integrated into the optimization of these volumetric representations for remote sensing applications.

\section{Method}

\subsection{Shadow-aware Earth Observation Dataset}

The dataset consists of 702 georeferenced tiles of $500 \times 500$~m each across the cities of Jacksonville, Omaha, and San Diego~\cite{CORE3D,dfc1,dfc2} as shown in Figure \ref{fig:dataset_overview}. For each tile, we provide multiple:
\begin{itemize}
    \item[--] Panchromatic (PAN) WorldView-3 images (30 cm)
    \item[--] Pansharpened RGB images (30 cm);
    \item[--] Bundle-adjusted RPC models for all PAN and pansharpened images;
    \item[--] DSMs obtained from LiDAR data from USGS with a 50~cm grid resolution. Two versions: min and max aggregation of altitude values;
    \item[--] Vegetation masks based on the  Normalized Difference Vegetation Index (NDVI);
    \item[--] Shadow masks. Two versions corresponding to min and max DSMs.
\end{itemize}

With multiple images per tile, captured on different dates and from various angles, we obtain a total of 19,162 images and masks of approximately $1500 \times 1500$ pixels. Note that the exact image size varies depending on the viewing angle.

\paragraph{Shadow mask generation.}
Manually annotating shadow masks is a time-consuming task that becomes infeasible at the scale of the dataset presented here. In~\cite{AISD}, the authors report having spent four months annotating just 514 images.

We use a shadow simulation algorithm to automatically generate shadow masks at scale from existing data. For each image in the dataset, the shadow mask is computed using the sun position at the time of capture, the aligned DSM for the area of interest, and the image camera models provided by the RPCs. Our algorithm consists of two steps: shadow casting, where shadows are simulated over the DSM based on the sun position and geometry, and shadow projection, where these shadows are projected onto each image using the corresponding RPCs. To make sure that the shadow projection step produces sufficiently dense maps, we generate higher resolution masks by upscaling the DSM by a factor of 4 before shadow casting.

The shadow casting algorithm simulates terrain shadows on the raster DSM by casting rays in the direction of the sun and marking occluded pixels as shadow (Algorithm~\ref{alg:shadowcast}). Discrete ray trajectories are computed for each pixel using a modified Bresenham algorithm~\cite{bresenham1998algorithm}. Along each ray, occlusion testing is performed using subpixel‐accurate sampling of the DSM and a threshold defined by the sun's elevation~ $\beta$.

Given the first pixel along a ray with elevation $Z_{\text{occluder}}$, by default this pixel is considered illuminated and defined as the first occluder. For each subsequent pixel along the ray with elevation \(Z_{\text{current}}\) and located at a horizontal distance \(d\) from the occluder, the horizontal length of the shadow casted by the occluder is 
\begin{equation}
    l = \frac{ Z_{\text{occluder}} - Z_{\text{current}}}{\tan{\beta}}.
    \label{eq:shadow_cast_eq}
\end{equation}
If the horizontal distance between the pixels is smaller than the shadow length $d < l$:
the current pixel is marked as a shadow; otherwise the pixel is considered illuminated and becomes the new occluder. The result is a shadow cast map that we denote $S_{DSM}$.

\begin{algorithm}[t]
\caption{Shadow casting algorithm}
\label{alg:shadowcast}
\begin{algorithmic}[1]
\Require A $\mathrm{DSM}$, \ie an altitude map of size $H \times W$, and the Sun position given by the azimuth $\alpha$ and elevation $\beta$ angles.
\Ensure A shadow mask \(S_{\mathrm{DSM}}\in\{0,1\}^{H\times W}\), where \(1\) indicates a shadow.
\State \(p \gets -\sin(\alpha) \times \cos(\beta)\) \Comment{x-component of sun dir.}
\State \(q \gets \cos(\alpha) \times \cos(\beta)\) \Comment{y-component of sun dir.}
\State \(a \gets \tan(\beta)\) \Comment{slope of sun rays}
\State \(M \gets 0\) \Comment{Initialize all pixels as illuminated}
\State \(\textit{Paths} \gets \text{ComputeBresenhamPaths}(W,H,p,q)\)
\For{\textbf{each} path \(\Pi\) in \(\textit{Paths}\)}
   \State \(\ell \gets \Pi[0]\) \Comment{Index of the first pixel of this path}
   \For{\(j = 1 \ldots |\Pi|-1\)}
      \State \((x_{\ell},y_{\ell}) \gets \text{SubPixelCoord}(\ell)\)
      \State \((x_j,\,y_j) \gets \text{SubPixelCoord}(\Pi[j])\)
      \State \(d \gets \sqrt{(x_j - x_{\ell})^2 + (y_j - y_{\ell})^2}\)
      \State \(Z_{\ell} \gets \text{BilinearInterp}(\mathrm{DSM},\,x_{\ell},\,y_{\ell})\)
      \State \(Z_j \gets \text{BilinearInterp}(\mathrm{DSM},\,x_j,\,y_j)\)
      \State \( l = ( Z_{\ell} - Z_j ) / a \) \Comment{Equation \eqref{eq:shadow_cast_eq}}
      \If{ $d < l$}
         \State \(M[\Pi[j]] \gets 1\) \Comment{Current pixel $\to$ shadow}
      \Else
         \State \(\ell \gets \Pi[j]\) \Comment{Current pixel $\to$ new occluder}
      \EndIf
   \EndFor
\EndFor
\end{algorithmic}
\end{algorithm}

After casting shadows on the DSM, the next step is to project them into the image coordinate system (Algorithm~\ref{alg:shadowproj}). The DSM pixel coordinates are localized to obtain the world coordinates, and then re-projected into the image coordinates using the RPC. Because of the projection, several DSM points may project onto the same image pixel. This is resolved with the use of a z-buffer, storing, for each image pixel, a pair $(Z_{\max}, S_{\max})$ containing the elevation of the highest projected DSM point $Z_{\max}$, and its corresponding value in $S_{DSM}$ \ie, $S_{\max}$. The shadow value for the image pixel is then given by $S_{\mathrm{img}} = S_{\max}$.

During the projection step, a small number of image pixels may not receive any corresponding projection from the DSM. These pixels are assigned a default non-shadow value and stored in an uncertainty mask to indicate areas without valid information. Finally, a post-processing step is applied to remove small spurious shadow regions using connected component analysis.

\begin{algorithm}[t]
\caption{Shadow projection from a DSM to an image}
\label{alg:shadowproj}
\begin{algorithmic}[1]
\Require A $\mathrm{DSM}$, a shadow mask $S_{\mathrm{DSM}}$, the image size $(H, W)$, RPC parameters associated to the image
\Ensure A projected shadow mask $S_{\mathrm{img}}$, an uncertainty mask $U$
\State $S_{\mathrm{img}} \gets \mathbf{0}^{H \times W}$ \Comment{Initialize with non-shadow}
\State $U \gets \mathbf{1}^{H \times W}$ \Comment{Initialize uncertainty mask}
\State $Z_{\text{buffer}} \gets -\infty^{H \times W}$ \Comment{Initialize Z-buffer}

\For{$(x,y) \in \mathrm{DSM}$}
    \State $(X,Y, Z) \gets \text{WorldCoords}(x,y)$ %
    \State $I \gets \text{RPCProjection}(X,Y,Z)$
    \If{$I$ is within image bounds \textbf{and} $Z > Z_{\text{buffer}}(I)$}
        \State $Z_{\text{buffer}}(I) \gets Z$ \Comment{Update Z-buffer}
        \State $S_{\mathrm{img}}(I) \gets S_{\mathrm{DSM}}(x,y)$ \Comment{Update shadow}
        \State $U(I) \gets 0$ \Comment{Mark as certain}
    \EndIf
\EndFor

\State $S_{\mathrm{img}} \gets \text{RemoveSmallRegions}(S_{\mathrm{img}})$
\end{algorithmic}
\end{algorithm}

While our shadow annotation method is well-suited for large-scale applications, it has certain limitations. First, DSMs and satellite images are often captured at different times—sometimes years apart—leading to discrepancies between the two (e.g., buildings appearing in one but not the other), which can result in missing or phantom shadows. Additionally, although our method primarily aims to label shadows of buildings and other static structures, transient objects such as large vehicles or seasonal vegetation changes can introduce spurious shadow artifacts. These artifacts are not always fully removed by post-processing or filtered out by vegetation masks.

Another limitation is the inability to handle hollow structures such as power lines or bridges. Since our method relies on elevation data to cast shadows, all objects are treated as solid walls, leading to inaccuracies in these cases. Furthermore, because our shadow masks are entirely derived from the DSM, any errors or inconsistencies in the DSM directly propagate into the generated shadows.

\begin{figure}[t]
    \hspace{-0.15cm}
    \includegraphics[width=\linewidth]{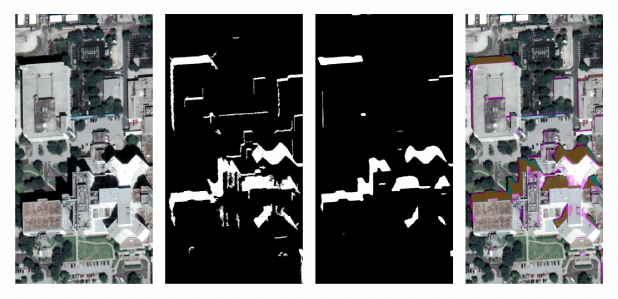}
    \caption{Limitations of our shadow annotation method. Left to right: pansharpened RGB image, shadow annotations, model predictions (\dataset{}-trained), and an overlay (magenta: ground truth, cyan: predictions, orange: matches). DSM holes in the cross-shaped building cause false positives, which the model corrects.}
    \label{fig:shadow_limitations}
\end{figure}

Figure \ref{fig:shadow_limitations} illustrates some of these challenges, along with predictions from a model trained on \dataset{}. Despite the noise in annotations, as detailed in the experiments section, the scale and overall quality of our dataset enable the training of robust models that ultimately generate even more accurate predictions than the provided labels.

\paragraph{Satellite data processing pipeline.}
The \dataset{} dataset primarily relies on two data sources: WorldView-3 imagery and corresponding metadata collected as part of the IARPA CORE3D program~\cite{CORE3D, dfc1, dfc2}, and LiDAR-derived elevation and terrain data from the USGS 3D Elevation Program.
The IARPA CORE3D data include respectively 26, 43 and 35 WorldView-3 images over Jacksonville, Omaha, and San Diego, captured between 2014 and 2016. Each WorldView-3 product contains panchromatic (PAN) images at 30~cm resolution and the corresponding multi-spectral (MSI) bands at 120~cm.

The \dataset{} processing pipeline begins with the definition of  $500 \times 500$~m square tiles, covering the footprints of all available images in each city. Tiles with more than $60\%$ water coverage, determined using the SRTM Water Body Data, are discarded. For each remaining tile, we extract the corresponding MSI and PAN crops from all images. The PAN images undergo radiometric correction. The MSI images first undergo a top-of-atmosphere correction and then are pansharpened. For that, each band is first independently aligned to the PAN using ORB feature matching~\cite{orb}.
Lastly, a vegetation mask is computed by thresholding the NDVI index at 0.0, allowing the removal of vegetation and tree shadows from further analyses.

To generate the DSMs, we use LiDAR data from the USGS 3D Elevation Program, primarily relying on the \textit{FL\_Peninsular\_FDEM\_Duval\_2018}, \textit{IA\_FullState}, and \textit{CA\_SanDiegoQL2\_2014} datasets. When data from these campaigns is unavailable, we use the most recent data available for the respective region. DSMs are computed at a grid resolution of 50 cm per pixel. 

To minimize the inclusion of shadows casted by thin or transient structures such as trees, light poles, and transmission lines, we aggregate the point cloud using the minimum elevation value in each cell (DSM Min). However, experiments (Section \ref{sec:ourResults}) revealed that DSM Min introduces a systematic bias, causing buildings to appear thinner and their shadows smaller than expected. To address this issue, we also compute a maximum-based DSM (DSM Max) and generate an alternative set of shadow masks (see Figure \ref{fig:dataset_overview}).

Lastly, we ensure precise alignment and georeferencing between the images and the DSMs. Our approach follows the same conceptual pipeline as~\cite{groundtruthGeneration}, but differs in specific implementation details, such as the choice of stereo models. First, we perform bundle adjustment~\cite{bundleAdjustment} on the PAN images to correct localization errors and discrepancies across the RPCs of different cameras. This step ensures internal consistency among cameras but does not provide absolute georeferencing with respect to the DSMs. After bundle adjustment, we generate a robust photogrammetric DSM by computing the median of 10 pairwise DSMs obtained with S2P~\cite{s2phd}. The image pairs for stereoscopic reconstruction are selected based on the heuristics of~\cite{gomezPairSelection,facciolo2017mvs}. Once the photogrammetric DSM is generated, we use it to estimate the shift between the images and the LiDAR data, ensuring full alignment between the ground-truth DSM and the S-EO imagery.

\subsection{Shadow Detection Network}
\label{sec:shadowDetectionNetwork}
We train a shadow detector for remote sensing images on the \dataset{} dataset, leveraging both the shadow masks and auxiliary masks. We experiment with two architectures: a {U-Net}~\cite{unet} and FSDNet \cite{fsdnet}, a fast, state-of-the-art, shadow detection model. To improve model generalization, we incorporate data augmentation strategies proposed in~\cite{yang2023silt}.

Our models are initialized with pretrained weights from a different domain (ground-level images). To better align our dataset’s color distribution—enhancing contrast and visual consistency—we apply a two-step color correction process. First, we perform a channel-wise quantile clip of intensity values~\cite{simplestColor} to align the color bands, followed by a histogram equalization~\cite{piecewiseHe} to enhance the contrast.

For training supervision, we employ the focal loss~\cite{lin2017focal} and explore different masking strategies. We exclude certain pixels from the loss computation using an uncertainty mask (marking pixels not projected from the DSM, see Algorithm~\ref{alg:shadowproj}) and a vegetation mask. 

\begin{figure}[t]
    \centering
    \includegraphics[width=0.95\linewidth, trim=50 60 5 150, clip]{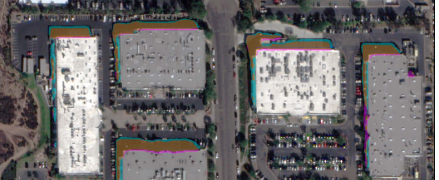}
    \caption{Impact of bias reduction when training with DSM Min and DSM Max shadow masks. The image shows shadow predictions for a San Diego tile: magenta (DSM Min-only training), cyan (Min-Max training), and orange (overlap, where both models agree). Magenta regions extend onto buildings, while cyan aligns better with actual shadows.}
    \label{fig:bias}
\end{figure}

\paragraph{Min-Max shadow masks.}
Our training strategy leverages the systematic differences between DSM Min and DSM Max. DSM Min tends to erode buildings, shrinking their structures and shadows, while DSM Max dilates them, expanding both. This creates a margin between the two, which we use to mitigate bias. As shown in Figure~\ref{fig:bias}, we supervise only in regions where both shadow masks agree, ignoring the rest. This reduces shadow overlap with buildings and improves the accuracy of predicted shadows (see Table~\ref{table:ourTable}).

\section{Experiments}

We conduct a series of shadow detection experiments to showcase the impact of our dataset and demonstrate its applicability beyond our evaluation setup. 
The \dataset{} dataset primarily functions as a large-scale training resource rather than a validation set, due to the inherent noise in its annotations (Figure~\ref{fig:shadow_limitations}). Our experiments show that it enables solutions to various shadow-related challenges across different domains.

\subsection{Shadow detection on the \dataset{} dataset}
\label{sec:ourResults}

During training, we randomly sample \(512 \times 512\) patches to generate training batches. We use a rolling window approach to ensure full coverage of validation images. We set the batch size to 32 and use a learning rate of \(5 \times 10^{-3}\) with the AdamWScheduleFree optimizer~\cite{schedulefree}. The training data consists of almost all Jacksonville sites, along with 50 sites from San Diego. The remaining data is reserved for validation and testing. The excluded Jacksonville sites correspond to tiles that Jacksonville sites JAX\_341, JAX\_342, JAX\_335, and JAX\_334 are also excluded from training as they overlap with EO-NeRF areas used later in Section~\ref{subsect-eonerf}.

Our initial experiments revealed no significant performance differences between FSDNet, initialized with weights from \cite{shadowsReview}, and a general-purpose U-Net pretrained on ImageNet. Given that U-Net is a widely used, well-understood architecture with broad applicability, we opted to conduct all subsequent experiments using the U-Net implementation from \cite{segmentationModels} with the usage of squeeze and excitation blocks \cite{squeezeAndExcite}.

\begin{figure}
    \centering
    \includegraphics[width=0.95\linewidth]{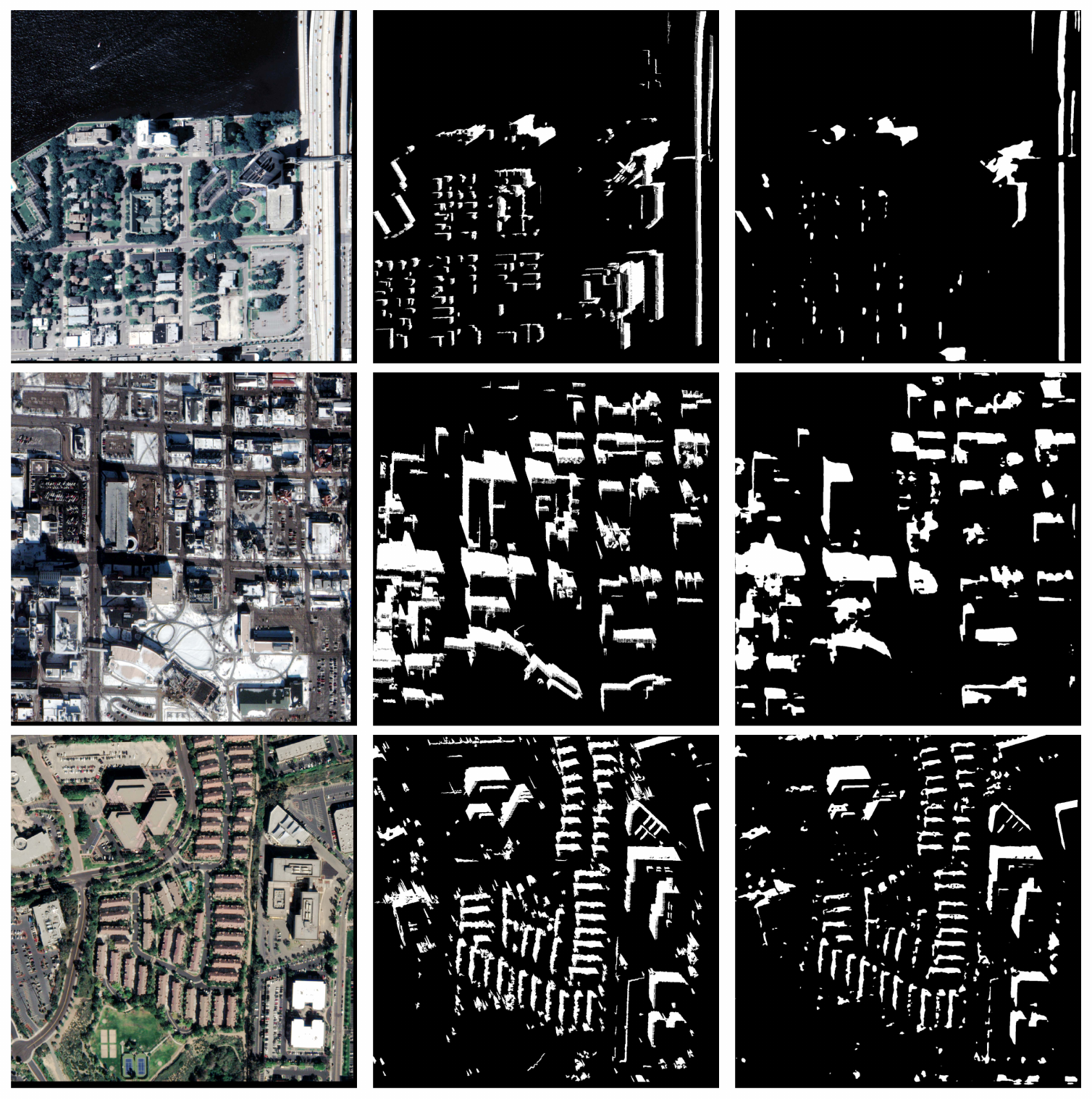}
    \caption{Qualitative results on the \dataset{} dataset. Top to bottom: Jacksonville (JAX\_334\_0), Omaha (OMA\_93\_15), San Diego (UCSD\_353\_10). Left to right: input image, ground-truth shadow mask, and model prediction.}
    \label{fig:our_preds}
\end{figure}

Table~\ref{table:ourTable} reports Balanced Error Rates (BER, Positive BER, Negative BER) and F-Score—standard shadow detection metrics~\cite{shadowsReview}—for both the Min-Max trained model and the baseline trained with Min-only masks. To mitigate annotation noise, we use uncertainty masks and evaluate under two conditions: (1) the full image and (2) only pixels that are certain and where both Min and Max shadow masks agree. We test on nine diverse areas with varied structures across the three cities, where we manually verified the accuracy of shadow annotations. Qualitative results are shown in Figure~\ref{fig:our_preds}.

\begin{table}[t]
    \centering
    \small{
    \setlength{\tabcolsep}{4pt} %
    \renewcommand{\arraystretch}{1.1} %
    \begin{tabular}{lcc|cc}
        \toprule
        \textbf{Metric} & \multicolumn{2}{c}{\textbf{Shadow detector}} & \multicolumn{2}{c}{\textbf{Baseline (min shadows)}} \\
        & \textbf{All} & \textbf{Filt.} & \textbf{All} & \textbf{Filt.} \\
        \midrule
        BER ($\downarrow$) & 30.81\% & 21.79\% & 33.59\% & 28.54\% \\
        Pos. BER ($\downarrow$) & 59.76\% & 43.17\% & 66.47\% & 56.86\% \\
        Neg. BER ($\downarrow$) & 1.86\% & 0.42\% & 0.71\% & 0.22\% \\
        F-Score ($\uparrow$) & 48.47\% & 70.13\% & 46.76\% & 59.13\% \\
        \bottomrule
    \end{tabular}}
    \caption{Evaluation metrics for shadow detection across diverse areas of the \dataset{} dataset: JAX\_341, JAX\_342, JAX\_335, JAX\_334, UCSD\_353, UCSD\_573, OMA\_93, OMA\_930, OMA\_967. Shadow detector is the final model trained with the min-max masks, baseline is only supervised with min shadows (Section \ref{sec:shadowDetectionNetwork})}
    \label{table:ourTable}
\end{table}

\begin{figure}
    \centering
    \includegraphics[width=0.95\linewidth]{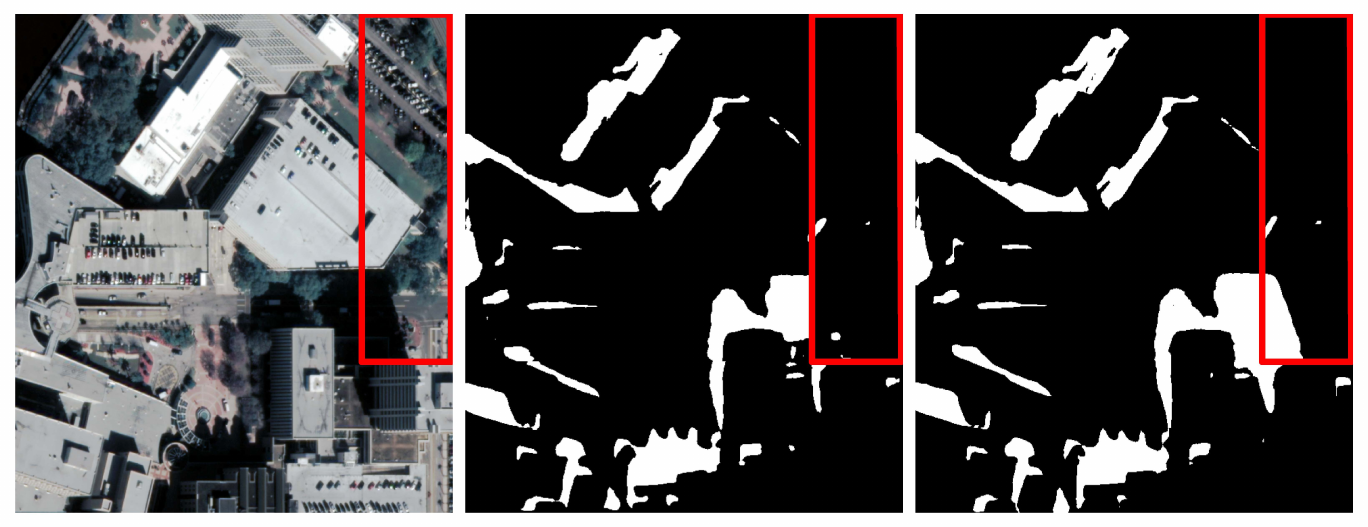}
    \caption{Impact of contextual information on shadow prediction. Left to right: input image, shadow mask from crop-based inference, and shadow mask from full-image inference. The red region highlights a crop with a misclassified shadow due to missing context in the cropped input.}
\label{fig:context_aware_shadows}
\end{figure}

Thanks to the use of fully convolutional models, we can process entire images at test time, allowing predictions to be context-aware. As shown in Figure \ref{fig:context_aware_shadows}, having access to the full scene can significantly impact shadow detection. When only a patch is provided, a dark region may not be labeled as a shadow if the object casting it is not visible. However, feeding the entire image into the model allows it to correctly associate shadows with their sources, demonstrating its ability to leverage contextual cues beyond pixel intensity.

\paragraph{Shadow detection generalization to AISD.}
We assess the generalization ability of our shadow detector on unseen remote sensing imagery. To this end, we use the AISD dataset~\cite{AISD}, which, to our knowledge, is the only available for this task. A noticeable domain shift exists between \dataset{} and AISD due to radiometric differences, as AISD is derived from aerial imagery. These discrepancies persist even after applying our color correction pre-processing, leading to suboptimal zero-shot generalization.

To bridge the gap between satellite and aerial domains, we fine-tune our model on a small subset of AISD’s training data, using only $10\%$ (41 images). Despite the limited supervision, this adaptation significantly enhances performance. Our fine-tuned model surpasses the best model reported in the benchmark \cite{AISD} in AUC-ROC and achieves a comparable F-score, as shown in Table~\ref{table:aisdTable}. 

\begin{figure}
    \centering
    \includegraphics[width=0.95\linewidth]{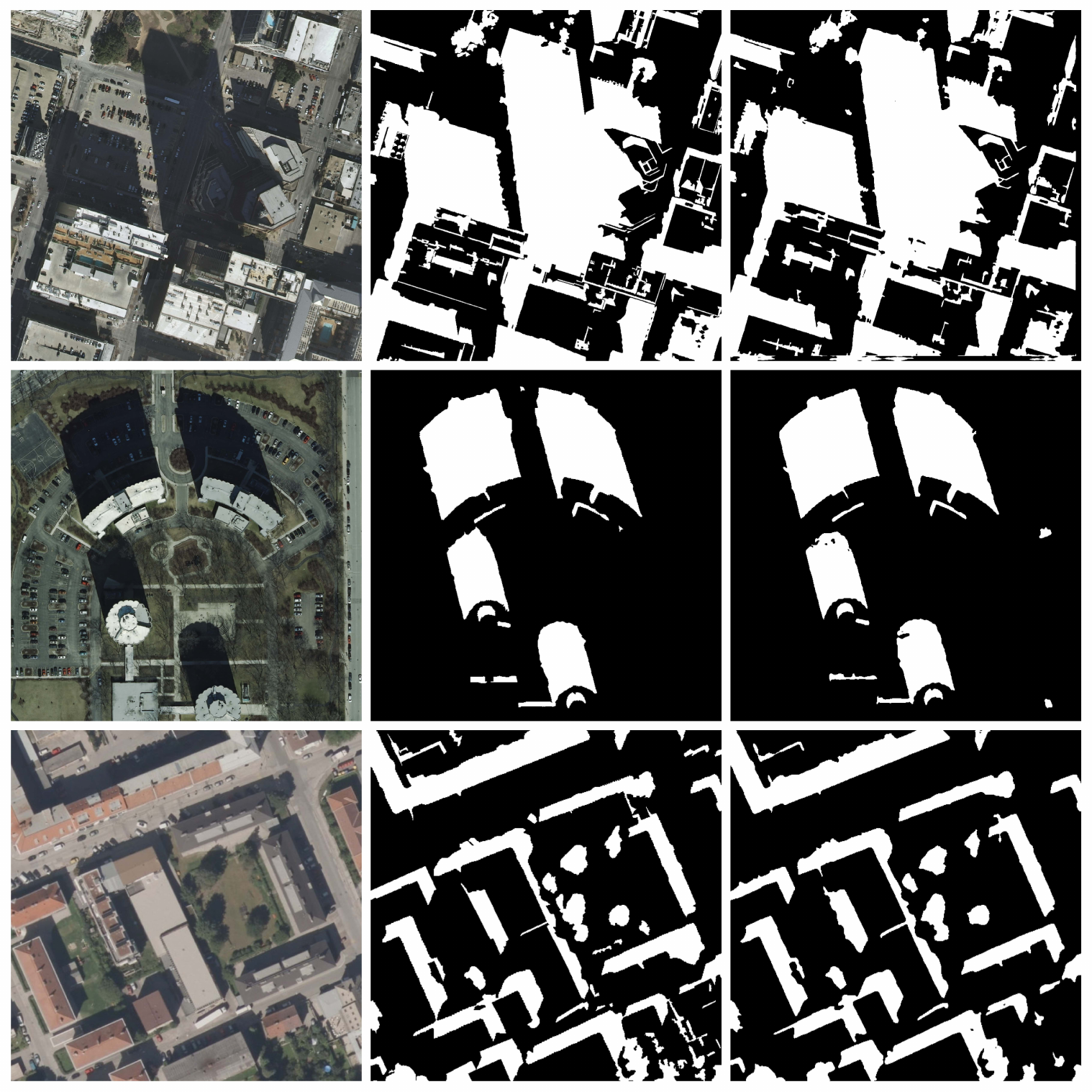}
    \caption{Qualitative results on the AISD dataset. Top to bottom: images from Austin, Chicago, and Innsbruck. Left to right: input image, ground-truth shadow mask, and model prediction.}
    \label{fig:aisd_preds}
\end{figure}

These results highlight the advantages of leveraging a large-scale pretraining dataset. To further validate this, we also train a U-Net from scratch using the full AISD training partition and find that our pretrained model outperforms it. This demonstrates that, in this case, pretraining on a large-scale dataset is more effective than training solely on task-specific data, even when the whole, but small, target dataset is available.
Moreover, our findings highlight the U-Net’s strong capability for shadow detection, showing that a it can surpass concurrent task-specific architectures when provided with sufficient data.  Table~\ref{table:aisdTable} reports the quantitative evaluation, and Figure~\ref{fig:aisd_preds} presents some visual results.

\begin{table}
    \centering
    \caption{Comparison of shadow detection performance on the AISD dataset. Higher values indicate better performance.}
    \label{table:aisdTable}
    \begin{tabular}{lcc}
        \toprule
        \textbf{Model} & \textbf{AUC-ROC} & \textbf{F-Score} \\
        \midrule
        DSSDNet \cite{AISD} & $0.985$ & $\mathbf{91.79\%}$ \\
        Fine-tuned U-Net (Ours) & $\mathbf{0.990}$ & $91.52\%$ \\
        U-Net from scratch & $0.976$ & $89.02\%$ \\
        \bottomrule
    \end{tabular}
\end{table}

\subsection{Extending EO-NeRF with shadow supervision}
\label{subsect-eonerf}

Lastly, we extended the {EO-NeRF} framework~\cite{eonerf} to use the shadow masks predicted by our segmentation model as auxiliary inputs, alongside the satellite views.

EO-NeRF is a NeRF variant designed for multi-date satellite imagery that achieves state-of-the-art geometry estimates thanks to a geometrically consistent shadow rendering approach. The EO-NeRF geometry is optimized by penalizing pixel-wise color differences between sets of rendered and actual pixels in the input views. Notably, the rendered color results from a combination of physical variables, including a shadow component. As a result, shadow masks can be easily integrated into the optimization process to help the shadow component align with the input masks.

Given the shadow values estimated by EO-NeRF, denoted $\hat{\mathcal{M}}$, and the GT shadow values in the input masks, denoted $\mathcal{M}$, we add the following term to the loss function:
\vspace{-0.5em}
\begin{equation}
    \mathcal{L}_S = \lambda \mathcal{M} ( \hat{\mathcal{M}} - \mathcal{M} )^2,
    \label{eq:shadow_loss}
\end{equation}
where $\lambda$ is a weight that we set to $\sum^N \! \mathcal{M}/N$, \ie , the percentage of GT shadow pixels in a batch of $N$ rays. Note that \eqref{eq:shadow_loss} penalizes only false negatives, \ie, rays not rendered as shadows that are shadows in $\mathcal{M}$. We also experimented with a binary cross-entropy term to penalize both false negatives and false positives, but it yielded poorer performance. Allowing the flexibility to incorporate new shadows, even those absent from the ground-truth mask, resulted in the best performance.

\begin{figure}
    \centering
    \includegraphics[width=0.99\linewidth]{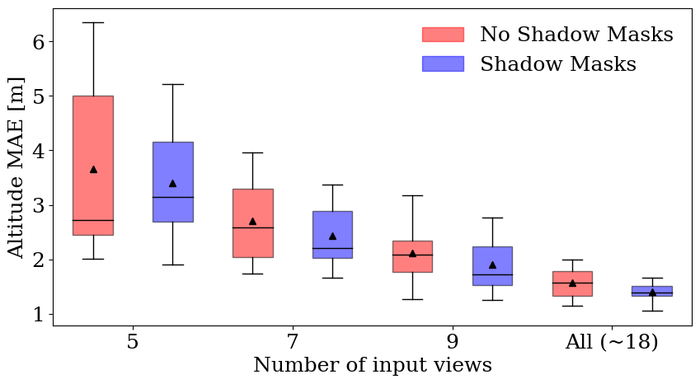}
    \vspace{0.15cm}
    
    \small{
    \begin{tabular}{@{\hskip 0.05cm}r@{\hskip 0.2cm}|c@{\hskip 0.25cm}c@{\hskip 0.25cm}c@{\hskip 0.25cm}c@{\hskip 0.25cm}}
    \hline
        \footnotesize{\bf{Altitude MAE [m]}} & \small{all views} & \small{9 views} & \small{7 views} & \small{5 views} \\
        \hline
        \small{Shadow masks, $\mathcal{L}_S$ \eqref{eq:shadow_loss}} & 1.40 & 1.90 & 2.44 & 3.41 \\
        \small{No shadow masks} & 1.57 & 2.11 & 2.71 & 3.66  \\   
    \end{tabular}}
    \caption{Average EO-NeRF altitude MAE (in meters) across all areas as a function of the number of input satellite views. Shadow supervision consistently improves altitude accuracy, as the error distribution of blue boxes is lower than that of the corresponding red boxes. Average MAE values are listed in the associated table.}
    \label{fig:eonerf_plot}
\end{figure}

\begin{figure}
        \hspace{-0.25cm}
        \begin{tabular}{@{\hskip 0.9cm}c@{\hskip 1.2cm}c@{\hskip 0.9cm}c@{\hskip 0.4cm}}
              Without $\mathcal{L}_S$ & With $\mathcal{L}_S$ & Ground-truth \\
             \multicolumn{3}{c}{   \includegraphics[width=0.99\linewidth]{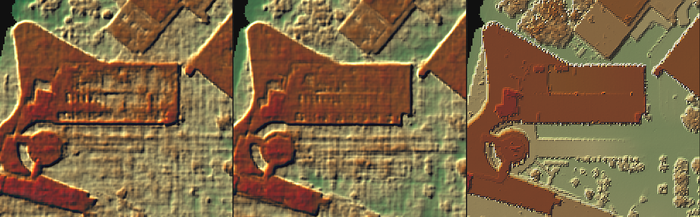}} 
        \end{tabular}
    \caption{Left to right: JAX\_214, 7 input views: EO-NeRF DSM detail without and with shadow supervision vs. LiDAR DSM.}
    \label{fig:eonerf_geometry}
\end{figure}

Figure~\ref{fig:eonerf_plot} shows the evolution of the altitude MAE with respect to the number of views across the \textit{DFC2019} and \textit{IARPA2016} areas of interest reported in EO-NeRF~\cite{eonerf}, with and without shadow mask supervision. In all experiments we used the bundle-adjusted RPC models provided in~\cite{mari2022satnerf, eonerf}. No view selection was performed; for each $K$-views subset, the first $K$ images from the training list were used. Experiments with fewer than 5 views are not included, as few-shot NeRF-based geometry becomes excessively noisy in the absence of depth priors~\cite{zhang2023spsnerf}.

The experiments that account for shadow masks using the loss \eqref{eq:shadow_loss} consistently achieve lower altitude MAE (Figure~\ref{fig:eonerf_plot}) and produce sharper, less noisy geometry compared to the concurrent experiments. Figure~\ref{fig:eonerf_geometry} shows a detail of the output geometry for one of the target areas~\cite{eonerf}.

\section{Conclusion}

This work presented \dataset{}, a novel large-scale, high-resolution dataset for geometry-aware shadow detection in satellite imagery. The dataset comprises WorldView-3 images, shadow masks, and ground-truth DSMs. All shadow masks in the dataset were automatically annotated using the sun position and the geometry of the associated DSMs. 

The significant potential of the \dataset{} dataset is showcased by training a shadow detector model and demonstrating its capacity to generalize to unseen aerial imagery. The predicted shadow masks are subsequently used to incorporate shadow supervision into the state-of-the-art multi-view reconstruction method EO-NeRF. The geometry estimates from the shadow-supervised variant consistently outperform the geometry reconstructions of the original method. Note that the proposed shadow-based supervision is also compatible with any other 3D modeling method that incorporates shadow modeling, including recent advancements in Gaussian Splatting~\cite{aira2024eogs}.

\subsubsection*{Acknowledgments}
This work was partially supported by Agencia Nacional de Investigación e Innovación (ANII, Uruguay) under the graduate scholarship POS\_NAC\_2023\_1\_177798. It was performed using HPC resources from GENCI-IDRIS (grant 2024-AD011012453R4). Centre Borelli is also with Universit\'{e} Paris Cit\'{e}, SSA and INSERM.

{
    \small
    \bibliographystyle{ieeenat_fullname}
    \bibliography{main}
}

\end{document}